\documentclass[10pt,twocolumn,letterpaper]{article}

\usepackage[pagenumbers]{cvpr} 

\definecolor{cvprblue}{rgb}{0.21,0.49,0.74}
\usepackage[pagebackref,breaklinks,colorlinks,allcolors=cvprblue]{hyperref}
\usepackage{amsmath}
\usepackage{pifont}

\def\paperID{ } 
\def\confName{CVPR}
\def\confYear{2026}

\title{CME-CAD: Heterogeneous Collaborative Multi-Expert Reinforcement Learning for CAD Code Generation}

\author{
Ke Niu\textsuperscript{1}\dag, 
Haiyang Yu\textsuperscript{1,2}\dag$^\ast$, 
Zhuofan Chen\textsuperscript{1}\dag,
Zhengtao Yao\textsuperscript{2}, 
Weitao Jia\textsuperscript{2} \\
Xiaodong Ge\textsuperscript{1}, 
Jingqun Tang\textsuperscript{2}, 
Benlei Cui\textsuperscript{2}, 
Bin Li\textsuperscript{1}$^\ast$,
Xiangyang Xue\textsuperscript{1}
\thanks{Corresponding author: Haiyang Yu (hyyu20@fudan.edu.cn). \dag Equal contribution.} \\
\textsuperscript{1}Fudan University, Shanghai, China. \\
\textsuperscript{2}ByteDance Inc. \\
{\tt\small \{kniu22, zfchen23, xdge25\}@m.fudan.edu.cn}, {\tt\small  zyao9248@usc.edu} \\ 
{\tt\small \{hyyu20, libin, xyxue\}@fudan.edu.cn}, {\tt\small tangjingqun@bytedance.com} \\
}

\begin{document}
\maketitle
\begin{abstract}
Computer-Aided Design (CAD) is essential in industrial design, but the complexity of traditional CAD modeling and workflows presents significant challenges for automating the generation of high-precision, editable CAD models. Existing methods that reconstruct 3D models from sketches often produce non-editable and approximate models that fall short of meeting the stringent requirements for precision and editability in industrial design. 
Moreover, the reliance on text or image-based inputs often requires significant manual annotation, limiting their scalability and applicability in industrial settings. To overcome these challenges, we propose the Heterogeneous Collaborative Multi-Expert Reinforcement Learning (CME-CAD) paradigm, a novel training paradigm for CAD code generation. Our approach integrates the complementary strengths of these models, facilitating collaborative learning and improving the model’s ability to generate accurate, constraint-compatible, and fully editable CAD models. We introduce a two-stage training process: Multi-Expert Fine-Tuning (MEFT), and Multi-Expert Reinforcement Learning (MERL). Additionally, we present CADExpert, an open-source benchmark consisting of 17,299 instances, including orthographic projections with precise dimension annotations, expert-generated Chain-of-Thought (CoT) processes, executable CADQuery code, and rendered 3D models. 
\end{abstract}

\section{Introduction}
\label{sec:intro}
As a fundamental tool in industrial design, Computer-Aided Design (CAD) has long been essential for the digital creation of complex engineering systems through its formalized modeling language. In modern smart manufacturing, the ``digital-first'' paradigm governs product development, where design concepts are translated into high-fidelity CAD models prior to physical production. However, this process is heavily reliant on engineers' specialized skills, which are honed through years of training. As a result, automating the generation of precise, constraint-compatible, and fully editable CAD models remains a significant challenge, and represents a critical area of focus for both industry and academia in the advancement of intelligent design.
\begin{figure}[t]
    \centering
    \includegraphics[width=1\linewidth]{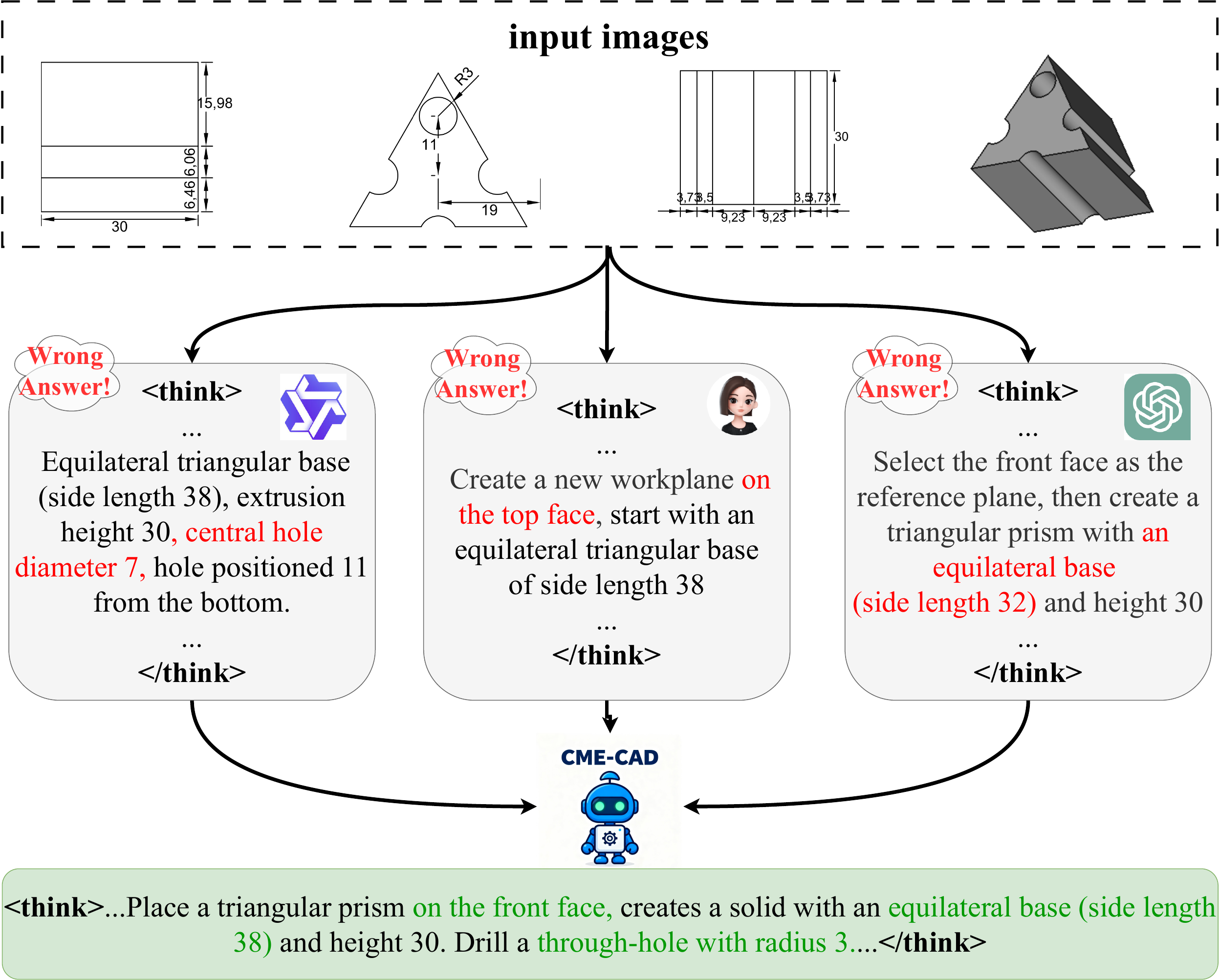}
\caption{The workflow of our method for CAD code generation.
}
\label{1}
\end{figure}

However, existing approaches to CAD code generation face two primary challenges. From the perspective of industrial design workflows, the current paradigm of generating CAD code from text and image inputs is both costly and difficult to implement in practical applications. This challenge arises primarily due to the necessity of manually annotating textual descriptions, which requires significant expertise. In contrast, 2D engineering drawings are more readily accessible and inherently contain the necessary information, based on design specifications, to construct accurate 3D models.
From a methodological standpoint, existing approaches predominantly rely on reinforcement learning with verifiable rewards (RLVR). However, because RLVR is an on-policy method, performance improvements are primarily achieved by optimizing the model along pre-existing, reward-rich reasoning paths. In essence, RLVR refines the model within its current knowledge base, rather than actively exploring novel information. When the reasoning paths are biased the model's ability is inherently restricted. Furthermore, in complex reasoning scenarios, the model’s initial strategy often fails to generate correct outputs, making the optimization process difficult. Therefore, it is essential to encourage the generation of more diverse responses within logically consistent reasoning paths, thereby increasing the likelihood of obtaining correct rewards.

To tackle these challenges, we focus on addressing two key aspects. First, from the perspective of industrial design workflows, we investigate the generation of precise CADQuery code directly from orthographic projection inputs, aligning more closely with real-world design practices. Second, from a methodological standpoint, inspired by the principle of ``learning from the strengths of others'' we introduce the Heterogeneous \textbf{C}ollaborative \textbf{M}ulti-\textbf{E}xpert Reinforcement Learning (CME-CAD) paradigm, specifically designed for generating executable, high-precision CAD code. The workflow of our method is shown in Fig.~\ref{1}. Central to this approach is the integration of multiple heterogeneous pre-trained models, which leverage their complementary strengths to facilitate collaborative learning. This collaborative learning process enhances the model's ability to generate accurate outputs, providing more reliable learning signals throughout training. Importantly, rather than simply aggregating the outputs of these models, we enable them to learn from one another’s strengths, fostering collaboration and driving the model’s overall performance to new heights. The training process consists of two stages:

\begin{itemize}
\item \textbf{Multi-Expert Fine-Tuning (MEFT)} focuses on generating reasoning paths with distinct styles by leveraging the unique capabilities of multiple heterogeneous experts. This stage refines each expert's individual reasoning pattern, enhancing their contributions to the overall model.

\item \textbf{Multi-Expert Reinforcement Learning (MERL)} promotes cross-expert learning by enabling knowledge transfer between experts. Additionally, it incorporates the hard negative sample buffering mechanism, ensuring the model continues to learn from difficult cases, mitigating reward sparsity, and significantly improving performance.

\end{itemize}

Another critical bottleneck in CAD code generation is the lack of high-quality open-source datasets that meet industrial-grade requirements. To address this, we introduce CADExpert, a benchmark specifically crafted for the generation of executable and editable CAD code. We have carefully designed a professional and robust method for dataset construction, ensuring that it aligns with real-world industrial needs. CADExpert contains 17,299 instances.
Each sample consists of four components: (1) orthographic projections with precise dimensional annotations, (2) detailed code generation processes produced by various expert models, (3) the corresponding executable CADQuery code, and (4) the 3D CAD model. This comprehensive dataset is designed to support robust training and evaluation of CAD code generation methods, providing a diverse and high-quality resource for advancing research in this domain.
Our contributions are as follows:
\begin{itemize}

\item We introduce the CME-CAD paradigm, a novel reinforcement learning framework for CAD code generation. This approach integrates the complementary strengths of multiple models, fostering collaborative learning and enhancing the model’s ability.

\item We present CADExpert, an open-source, industrial-grade dataset that includes: (1) three-view images with precise dimension annotations, (2) code generation processes (COT) from various experts, (3) corresponding executable CADQuery code, and (4) rendered 3D CAD models.

\item  By aligning the problem with real-world workflows, we bridge the gap between academia and industry and conduct extensive evaluations of state-of-the-art Vision-Language Models (VLMs) in this context.
\end{itemize}

\section{Related Work}
\label{sec:formatting}

\subsection{CAD Code Generation}

CAD Code Generation aims to directly generate precise and editable 3D models based on user instructions. Compared to general code generation tasks, CAD code generation presents unique challenges, such as the high demand for model reasoning capabilities. This includes the need to understand spatial, geometric, and physical properties, as well as being sensitive to numerical errors.
Approaches like CAD-MLLM~\cite{xu2024cad} and GenCAD~\cite{alam2024gencad} utilize vision-language models to map image inputs to CAD commands, enabling direct translation from visual representations to design instructions. Img2CAD~\cite{you2024img2cad} adopts a two-stage design that separates structure prediction from parameter regression, improving the flexibility of the model. CAD2Program~\cite{wang20252d} introduces a more flexible code format that enhances the expressive power of the model, facilitating more complex and richer modeling. CADCodeVerify~\cite{alrashedy2024generating} proposes an iterative verification and improvement approach for CAD code generation, refining generated code through continuous feedback loops. CAD-Llama~\cite{li2025cad} develops a hierarchical annotation pipeline that transforms parameterized 3D CAD command sequences into structured parametric CAD code. In a similar vein, CAD-Coder~\cite{guan2025cad} achieves high accuracy by generating structured CadQuery code from expert-written descriptions, significantly improving the reliability and precision of generated CAD models. CAD-RL~\cite{niu2025intent} is the first to successfully generate numerically accurate CADQuery code for CAD model creation.

\subsection{Reinforcement Learning with Verifiable Rewards}
With the development of Vision-Language Models (VLMs) ~\cite{yu2025benchmarking,yu2025eve,yu2025umit,niu2025chatreid,niu2025creft, yin2025sail},Reinforcement learning with verifiable rewards (RLVR)~\cite{kojima2022large,wei2022chain,wang2022self} has demonstrated its effectiveness in facilitating complex reasoning tasks. However, despite significant improvements in performance through test-time computational scaling, their experiments reveal that the model's performance is still constrained by its inherent knowledge. Recent works by~\cite{ye2025limo,pang2025bolt,fatemi2025concise,zhao2025absolute} highlight that introducing structured Chain-of-Thought (CoT) reasoning paths can significantly enhance the model's reasoning capabilities. Additionally,~\cite{su2025trust,wen2025light,yang2025thinking} have introduced novel training paradigms designed specifically to improve reasoning.
Nevertheless, recent studies~\cite{shojaee2025illusion,zhao2025echo} indicate that the on-policy learning nature of RLVR encounters difficulties in exploring the reasoning space. This limitation arises because RLVR typically biases the model toward actions more likely to yield rewards, rather than encouraging the exploration of novel knowledge or new reasoning paths. As a result, these methods tend to rely on familiar reasoning patterns rather than discovering new, knowledge-based reasoning approaches. To address this challenge, we propose the heterogeneous Collaborative Multi-Expert Reinforcement Learning (CME-CAD) paradigm, which leverages the diverse reasoning paths generated by heterogeneous pre-trained models. This approach not only overcomes cognitive constraints but also preserves the self-driven exploration capabilities of the model, enabling more robust and innovative reasoning.

\begin{figure*}[t]
    \centering
    \includegraphics[width=1\linewidth]{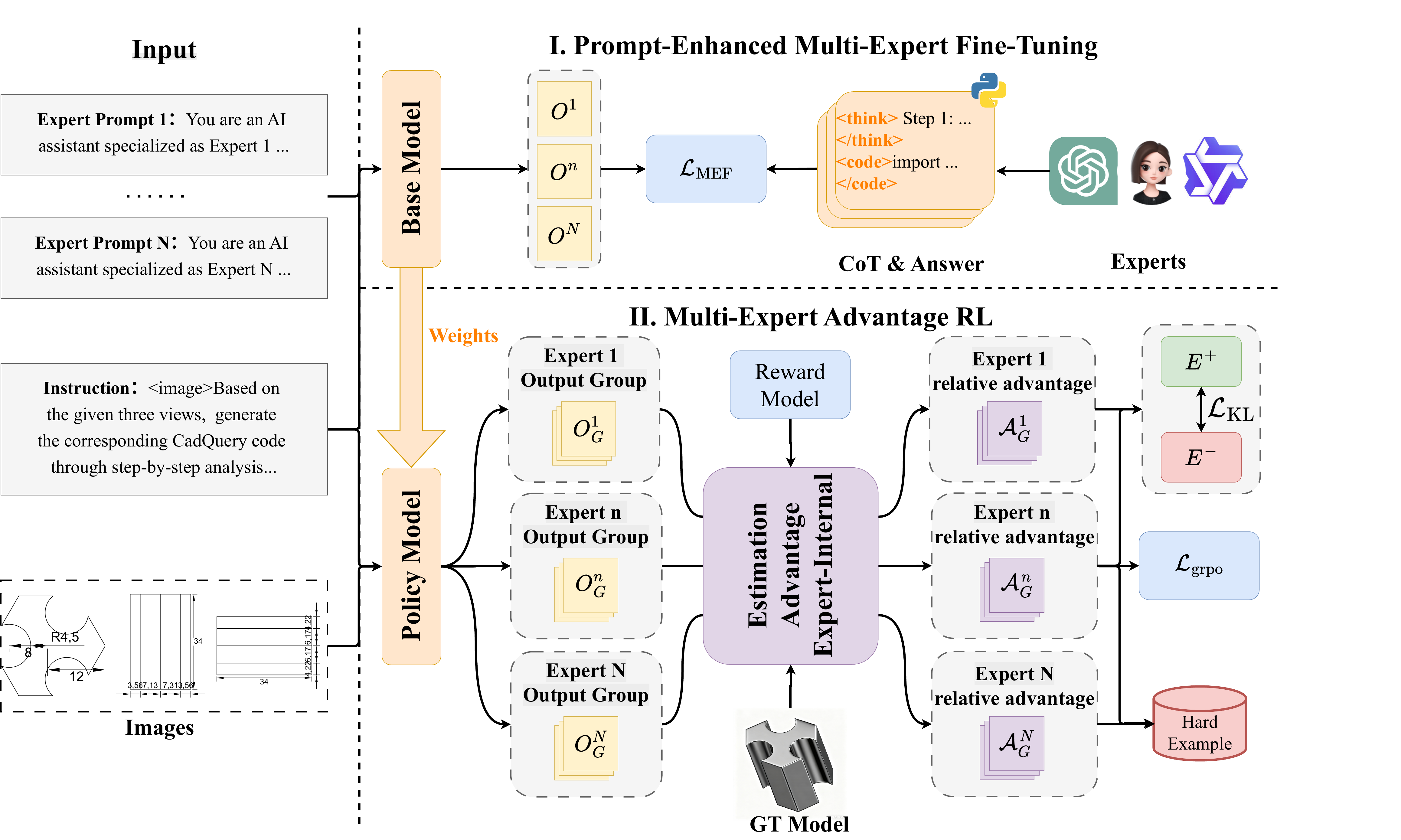}
\caption{Overall architecture of our method CME-CAD framework. $O$ represents the model output. $ O^N $ denotes the concatenation of the $ N $-th expert's Chain-of-Thought (CoT) and corresponding answer. $ O_G^N $ represents the group of $G$ outputs generated by the $ N $-th expert. }
\label{2}
\end{figure*}

\section{Method}

In this section, we detail the proposed Heterogeneous Collaborative Multi-Expert Reinforcement Learning (CME-CAD) framework. Fig.~\ref{1} illustrates the overall structure of our method. Sec.~\ref{3.2} presents the first training stage, Multi-Expert Fine-Tuning (MEFT). Sec.~\ref{3.3} elaborates on the second training stage, Multi-Expert Reinforcement Learning (MERL). Sec.~\ref{3.4} introduces our newly proposed dataset, CADExpert, designed specifically for executable and editable CAD code generation.

\subsection{Multi-Expert Fine-Tuning (MEFT)}
\label{3.2}

\textbf{Heterogeneous Multi-Expert Chain-of-Thought (CoT) Generation.} To construct expert-specific samples, we begin by selecting $N$ heterogeneous expert models, each distinguished by a unique system prompt $P_n$. We define the input set as 
$I$, which includes both the instructions and the 2D engineering drawings. For each pre-trained expert, we pose a instruction $I_i$ and obtain the corresponding answer $A_i^{(n)}$ along with the expert’s CoT $C_i^{(n)}$, which details the reasoning path. These outputs are then used to construct an expert-specific sample $S_n$, which encapsulates the distinct reasoning style and output of each expert.

\begin{equation}
S_n = \left( P_n, I_i, C^{(n)}_i, A^{(n)}_i \right).
\end{equation}

Existing leading pre-trained models face significant challenges in generating CADQuery code directly from 2D drawings. This difficulty arises primarily due to the lack of relevant pre-training data and tasks during pre-train phase. As a result, the reliability of the CoT generated for such tasks is limited. To address this issue, we propose a reverse-task approach. Specifically, we provide the model with orthographic projections alongside their corresponding CADQuery code. We guide the model to reason about how to derive the corresponding code from the orthographic projections, thereby generating a more reliable rasoning path.

\textbf{Training Process.} In this stage, the model is trained to predict the concatenated CoT $C^{(n)}_i$ and the corresponding final answer $A^{(n)}_i$, conditioned on the input $I_i$ and system prompt $P_n$. 
The training objective is to maximize the likelihood of jointly generating a coherent reasoning process and a correct final answer. This is achieved by minimizing the negative log-likelihood, which encourages the model to produce outputs that align with the expert's reasoning and answer. Here, $p(\cdot \mid \cdot)$ denotes the conditional probability mass function that models the likelihood of generating the target concatenated sequence given the input and system prompt. The objective function is:

\begin{equation}
\mathcal{L} = - \sum_{n=1}^{N} \sum_{i=1}^{I} \log \left( p(\text{Concat}(C^{(i)}_i, A^{(i)}_i) \mid P_n, I_i) \right).
\end{equation}

\subsection{Multi-Expert Reinforcement Learning (MERL)}
\label{3.3}
In this section, we provide a detailed overview of the second training stage, Multi-Expert Reinforcement Learning (MERL). We focus on key components such as the Reward Function Design, Expert-Internal Advantage Estimation, Multi-Expert Collaborative Learning, and the training mechanism, Hard Negative Sample Buffering Mechanism.

\textbf{Reward Function Design.} To ensure that the reward function is both robust and flexible, we align the model outputs with four key objectives. First, we enforce that the output format adheres to a specific structure, combining the expert model’s reasoning paths with the generated code. Second, we validate the executability of the generated code. Third, we assess the geometric consistency of the generated 3D model with the ground truth. Finally, we ensure that the coordinate system consistency is maintained throughout the generated model.

For the {Format Reward} $R_{\text{format}}$, we employ a regular expression matching mechanism to enforce the generation of a structured output by the model, where the reasoning process precedes the generated code. The reward is assigned a value of $1$ if the model generates the output in the correct format; otherwise, the reward is $0$.

The {Executability Reward} $R_{\text{exec}}$ evaluates whether the generated CAD code is both syntactically correct and executable within a Python environment. Since the model's output is CadQuery-based Python code, we use the Python interpreter to validate its syntax and check for runtime errors. If the code parses correctly and executes without errors, it receives a reward of $1$; otherwise, the reward is $0$.

The {Geometric Accuracy Reward} $R_{\text{IoU}}$ measures the geometric precision of the generated 3D model in comparison to the reference model. The generated code is executed to produce the 3D model, which is then compared to the ground-truth model using the Intersection-over-Union (IoU) metric. The IoU is calculated using the Jaccard Index $J$:

\begin{equation}
    R_{\text{IoU}}(M_{\text{gen}}, M_{\text{gt}}) = J(M_{\text{gen}}, M_{\text{gt}}) =  \frac{|M_{\text{gen}} \cap M_{\text{gt}}|}{|M_{\text{gen}} \cup M_{\text{gt}}|}.
\end{equation}

In 3D model reconstruction, the accuracy of the reference coordinate system is critical. During training, even if the model's reconstruction is geometrically correct, any deviation in the coordinate system can result in a complete failure of the IoU metric. To address this issue, we introduce the Work Plane Reward $R_{\text{plane}}$, where the uniqueness of the coordinate system is determined by both the position of the origin and the orientation of the coordinate axes. We quantify two key sources of error: the origin deviation $Dis_{\text{ori}}$ and the normal vector deviation $Dis_{\text{vec}}$.
The origin deviation is calculated as the Euclidean distance between the origin of the generated model and the true origin, defined as:

\begin{align}
    Dis_{\text{ori}} &= \| O_{\text{gen}} - O_{\text{gt}} \|_2 .
\end{align}

The normal vector deviation quantifies the misalignment of the model's coordinate axes. This is computed as:

\begin{equation}
    Dis_{\text{vec}} = \frac{1}{2} \left[ 2 - \text{sim}(\boldsymbol{x}_{\text{gen}}, \boldsymbol{x}_{\text{gt}}) - \text{sim}(\boldsymbol{y}_{\text{gen}}, \boldsymbol{y}_{\text{gt}}) \right].
\end{equation}

The overall coordinate system consistency reward $R_{\text{plane}}$ combines the two errors as follows:

\begin{equation}
R_{\text{plane}} = 1 - \beta \cdot Dis_{\text{ori}} - \gamma \cdot Dis_{\text{vec}}, \quad 0 \leq R_{\text{plane}} \leq 1.
\end{equation}

The total reward calculation is contingent upon satisfying both the {Format Reward} and the {Executability Reward} as essential conditions. Thus, we adopt a gating mechanism for these two components: the total reward can only be positive if both of these core conditions are met. The total reward function is defined as:

\begin{equation}
   R = \lambda_{\text{format}}R_{\text{format}} \cdot \lambda_{\text{exec}} R_{\text{exec}} \cdot\left( \lambda_\text{IoU} R_{\text{IoU}} + \lambda_{\text{plane}}R_{\text{plane}} \right) .
\end{equation}

\textbf{Expert-Internal Advantage Estimation.}
We use the model trained in the first stage as the current policy model. For each sample, it consists of {system prompts} $P_n$ and the input $I_i$. Different system prompts $P_n$ guide each expert's strategy, inducing the sampling of $G$ responses for each expert. Each response is generated by the current policy model parameterized by $\theta$.
Given $N$ experts, we generate $N \times G$ responses, where each response is denoted as $(C_g^n, A_g^n)$.
Each response $(C_g^n, A_g^n)$ is evaluated using a reward function $R$, which calculates an {absolute reward} $R_g^n$ for each response based on its quality:

\begin{equation}
R_g^n = R(C_g^n, A_g^n).
\end{equation}

Next, we calculate the relative advantage $\mathcal{A}_n$ within each expert's group of responses. The relative advantage for the $g$-th response of expert $n$ is computed by subtracting the average reward of all $G$ responses from the absolute reward of the $g$-th response:

\begin{equation}
\mathcal{A}_n = R_g^n - \frac{1}{G} \sum_{g'=1}^{G} R_{g'}^n.
\end{equation}

Thus, the relative advantage $\mathcal{A}_n$ quantifies how much better the $g$-th response is compared to the average of all responses generated by expert $n$. 
Based on the relative advantage $ \mathcal{A}$, we compute the GRPO~\cite{shao2024deepseekmath} loss function. It is important to note that we add a {non-negative truncation term} $ \max(A_g^n, 0)$ to avoid excessively penalizing the model's exploration ability. Given the complexity of the code generation task, truncating negative advantage values to zero ensures that the model is not immediately discouraged from exploring uncertain actions due to small negative advantages caused by minor errors in the process.
The {GRPO loss}  can be written as:

\begin{equation}
\mathcal{L}_\text{GRPO}^{(n)} = - \mathbb{E}_{A_g^n \sim \pi_{\theta}} \left[ \log \pi_{\theta}(A_g^n |  P_n, I_i) \cdot \max(\mathcal{A}_g^n, 0) \right].
\end{equation}

The term $ \pi_{\theta}(A_g^n |  P_n, I_i)$ represents the {probability} of generating the response $ A_g^n$ by expert $n$, given the input $I_i$ and system prompt $P_n$, under the current policy model parameterized by $ \theta$. The quantity $ \mathcal{A}_g^n$ denotes the {advantage} for the $g$-th response from expert $n$. The expression $ \max(\mathcal{A}_g^n, 0)$ is the {non-negative truncation} function.

\textbf{Multi-Expert Collaborative Learning} To facilitate knowledge exchange among experts, we introduce a multi-Expert collaborative learning mechanism. For each input $I_i$, we compute the {average absolute reward} $\overline{r_n}$ for expert $n$ across all generated responses. The expert with the highest average reward is defined as the {best expert} $E^{+}$, while the expert with the lowest average reward is defined as the {worst expert} $E^{-}$.
Thus, $\log p_{\theta}(A_{\text{correct}} | P_{E^{-}}, I_i)$ represents the log-probability of a high-quality response from $E^{-}$, and $\log p_{\theta}(A^+ | P_{E^{+}}, I_i)$ represents the log-probability of a high-quality response from $E^{+}$.
To promote learning from the more effective experts, we introduce a {KL divergence penalty}, which forces less effective experts to learn from the better ones.
We define the {KL divergence penalty} as follows:

\begin{equation}
\mathcal{L}_{\text{KL}} = \text{KL}\left( \pi_{\theta}(A^+ |  P_{E^{-}},I_i) \| \pi_{\theta}(A_{\text{correct}} |  P_{E^{+}},I_i) \right).
\end{equation}

Thus, the {KL divergence loss} ensures that the worse-performing expert $E^{-}$ learns from the more effective expert $E^{+}$ and gradually improves its performance. The KL divergence penalty is used to transfer knowledge from high-reward solutions, without reducing the model's diversity. 

This characteristic is achieved through the use of fixed and unique system prompts, which enforce differentiated reasoning styles across experts. The core objective is to guide the model to reason: \textit{``Given your unique prompt style, how would you generate this correct answer?''} Additionally, by leveraging the vLLM inference engine for parallel batch sampling (rollouts) and the paged attention mechanism, the total training time increases by only 20\%-30\% compared to the single-expert baseline, instead of the theoretical $N$-fold increase. Moreover, since vLLM pre-allocates memory based on the inference configuration, peak memory usage remains comparable. A key advantage during inference is the ability to select the best-performing expert, significantly reducing computational costs, as opposed to traditional multi-expert architectures, which require running all experts and selecting the optimal one while maintaining high-quality responses.

\begin{figure*}[t]
    \centering
    \includegraphics[width=0.9\linewidth]{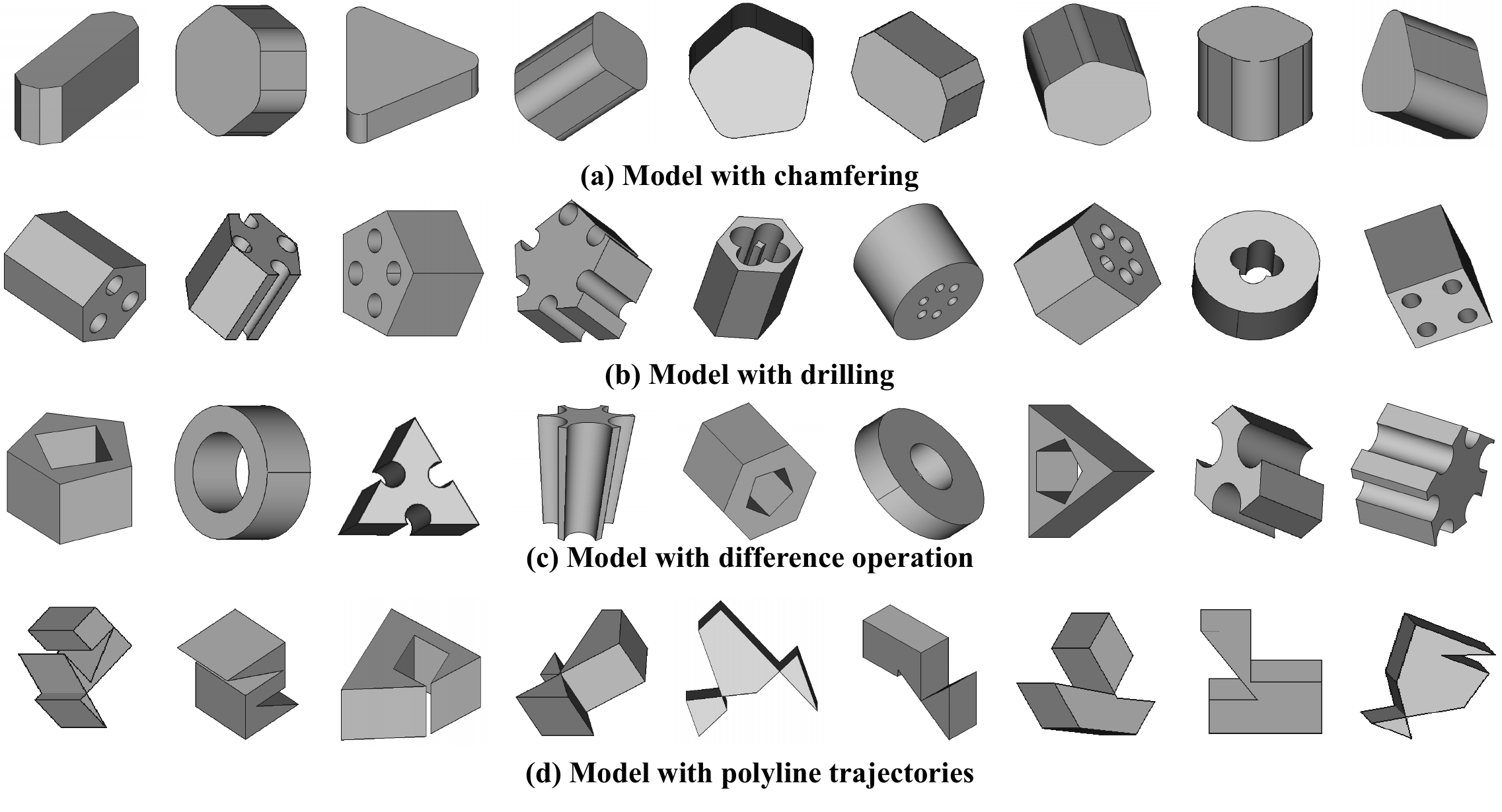}
\caption{Examples from CADExpert, showcasing more complex industrial design challenges, including intricate features that align with real-world modeling difficulties.}
\label{3}
\end{figure*}

\textbf{Hard Negative Sample Buffering Mechanism.} The Hard Negative Sample Buffering Mechanism addresses the issue when all experts fail to generate a correct response for a particular query.
We divide the reinforcement learning data into $M$ parts. After training on one part (i.e., $\frac{1}{M}$ of the data), we use the $\frac{1}{M}$ portion as the test set. We maintain a {buffer $B$} that stores difficult samples. For each input $I_i$, if expert $n$ generates more than $K$ incorrect answers out of $G$ responses, we calculate the probability of adding the sample to the buffer as $\frac{K}{G}$. 
Subsequently, we perform supervised fine-tuning on the data stored in the buffer $B$. The supervised loss for each input $I_i$ is calculated by comparing the model's output with the ground-truth answer.
Thus, the total supervised fine-tuning loss is:

\begin{equation}
\mathcal{L}_{\text{SFT}} = - \sum_{B} \log p_{\theta}(A_{\text{correct}} | P_n, I_i).
\end{equation}

 By revisiting these challenging samples and performing targeted fine-tuning, the model can learn to better handle a broader range of input scenarios. This mechanism ensures that even the most difficult examples are not discarded, thus improving the utilization of all available data. As a result, the model's robustness is enhanced, especially when dealing with edge cases or scenarios that are not adequately covered by simpler samples. This process promotes continuous improvement.

\subsection{CADExpert}
\label{3.4}

A significant bottleneck in the progress of CAD code generation lies in the lack of high-quality open-source datasets that meet industrial-grade standards. Existing datasets suffer from several key issues, including:
\begin{itemize}
    \item \textbf{Lack of Availability}: Descriptions of trajectories and other detailed textual data are often not available.
    \item \textbf{Missing Precise Dimensional Information}: The input images used for CAD generation are usually sketches without precise, detailed dimensional annotations.
    \item \textbf{Simplistic Operations}: Existing datasets lack examples of complex industrial operations, such as chamfering, drilling, and difference operations.
\end{itemize}

To address these challenges, we introduce CADExpert, a benchmark dataset specifically designed for executable and editable CAD code generation. CADExpert contains 17,299 instances. As shown in Fig.~\ref{3}, our model addresses several challenges not typically included in previous datasets, but which are essential for real-world applications. These challenges include operations such as chamfering, drilling, set difference, and complex trajectories, all of which reflect the intricacies encountered in industrial design tasks.
Each instance in the dataset consists of: orthographic projections with precise dimension annotations, detailed reasoning paths generated by different expert models, the corresponding executable CADQuery code, and the rendered 3D CAD model.

\textbf{Data Generation Process.}
The data generation process consists of several key steps:
\begin{itemize}
    \item \textbf{Random Generation of CADQuery Code with Constraints}: On random base planes (e.g., XY, YZ, XZ), we design multiple basic shape trajectories (e.g., rectangles, regular polygons, circles, polyline trajectories). These shapes are then extruded along a vertical plane and subjected to various complex operations, such as drilling, chamfering, and difference operations. The resulting CADQuery code is executable, editable, and randomly generated with specific constraints.
    \item \textbf{CADQuery Code Filtering}: After generating the random CADQuery code, we implement a comprehensive two-stage filtering process to ensure both the validity and functional integrity of the code.
In the first stage, executable validation is performed by running the generated code through a Python interpreter to check for syntax errors, runtime exceptions, and logical consistency. 
In the second stage, the manual review process is initiated. A team of 10 experts is responsible for performing a sampling-based review of the generated code. The reviewers visually inspect the generated CAD models to verify that they align with the intended design features, focusing on geometric accuracy and the correct implementation of the code’s functionality. In cases where ambiguities or uncertainties arise, the reviewers collaborate to reach a consensus. For particularly complex models, we implement a multi-round validation mechanism. In this process, the reviewers may iterate on the code generation process and provide feedback to refine the code, ensuring that the final output meets the required quality standards and adheres to the intended design specifications.

    \item \textbf{Generating Annotated Images}: Based on the generated CADQuery code, we execute the code to produce a STEP file. Using the FreeCAD~\cite{riegel2016freecad} API, we convert this STEP file into orthogonal projections and generate accurate 2D DXF files. These DXF files are then annotated with dimension lines using the \texttt{ezdxf} library, and the final DXF files are rendered into raster images.
    \item \textbf{Expert CoT Process Generation}: Using multiple expert models, we input the generated code and corresponding orthogonal projections to produce varying reasoning paths. Each expert generates a unique reasoning path for solving the problem, contributing to the diversity of the dataset.
\end{itemize}

This detailed process aims to resolve the major limitations of existing datasets by providing more accurate, complex, and industry-relevant data, thus facilitating more effective and robust CAD code generation models.

\section{Experimental Results}

\subsection{Implementation Details}

The base model used for this experiment is Qwen3-4B-Instruct~\cite{yang2025qwen3}, and the training is conducted on a setup with 8 H100 GPUs. During the Multi-Expert Fine-Tuning stage, the learning rate is set to $1 \times 10^{-5}$, with a batch size of 32. In the Multi-expert rainforcement learning stage, the learning rate remains at $1 \times 10^{-5}$, with a batch size of 8. For all RL-trained models, we maintain 4 rollouts per prompt. The temperature is set to 0.9 to control the randomness of predictions during training.
We utilize three expert models in our setup. {Expert 1} is {Qwen3–VL-Plus~\cite{yang2025qwen3}}, {Expert 2} is {GPT-5-Mini~\cite{achiam2023gpt}}, {Expert 3} is {Doubao-Seed-1.6-Vision~\cite{guo2025seed1}}.
For inference, to ensure consistent results and eliminate the impact of randomness, no sampling methods (such as temperature sampling or beam search) are employed during the testing phase for any of the models.

\subsection{Evaluation Metrics}
We evaluate the performance of Vision-Language Models (VLMs) and our method on CADExpert using four complementary metrics that focus on geometric accuracy, code executability, and structural precision. These metrics include:
(1) Intersection-over-Union (IoU): This metric measures the volumetric overlap between the predicted and reference shapes. Given its strict spatial alignment requirement, IoU is particularly effective for assessing the accuracy of shapes in industrial CAD applications. Higher value is better for IoU.
(2) Mean Chamfer Distance (Mean CD): This metric calculates the average geometric discrepancy between the generated and ground-truth models in point cloud space. It provides an overall measure of shape fidelity.
(3) Median Chamfer Distance (Med CD): This metric captures the typical error per sample, offering a more robust measure by being less sensitive to outliers. It provides a stable estimate of model performance. Lower values are preferred for CD.
(4) Executability (Exec.): This metric directly evaluates the functional validity of the generated code, assessing whether the generated CAD model can be executed correctly. Higher value is better for executability.

\begin{table}[t]
\centering
\small
\setlength{\tabcolsep}{2pt} 
\caption{The comparison on CADExpert is presented in two sections. The upper block reports results using pretrained SOTA VLMs without any fine-tuning, while the lower block displays results after fine-tuning. $\ast$ denote our re-implementation trained on the same benchmark.}
\label{result}
\begin{tabular}{l|cccc}
\toprule
 {\textbf{\textsc{Model}}} 
& IOU(\%)$\uparrow$ & Mean CD$\downarrow$  & Med CD$\downarrow$ &Exec.(\%)$\uparrow$ \\

  \midrule
LLaVA-1.5~\cite{liu2024improved} & 0.73 & 29.36 & 8.14 & 4.79\\
Phi-3.5-Vision~\cite{abdin2024phi} & 3.44& 27.92& 7.75 & 8.50\\
InternVL2.5~\cite{chen2024internvl} & 11.98 & 22.27& 7.36 & 23.92\\
Qwen2.5-VL~\cite{qwen2.5-VL} & 19.21& 20.68 & 6.64 & 31.27\\
InternVL3~\cite{zhu2025internvl3exploringadvancedtraining}& 24.59 & 21.40 & 6.62 & 29.68 \\
Gemini2.5 Pro~\cite{gemini2.5pro2024}& 30.86 & 14.13 & 5.97 & 39.40\\
GPT5-Mini~\cite{achiam2023gpt} & 35.15 & 9.89 & 4.81& 47.13\\
Doubao-1.6~\cite{guo2025seed1}& 35.62 & 7.54 & 4.35 & 52.89\\
Qwen3-VL~\cite{yang2025qwen3} & 37.04 & 6.96 & 3.84 & 54.79\\

\midrule
CAD-RL$\ast$ ~\cite{niu2025intent}& 71.84 & 1.38 & 0.36 & 97.32\\
Ours & \textbf{80.71} & \textbf{1.00} & \textbf{0.11} &  \textbf{98.25}   \\
\bottomrule
\end{tabular}
\end{table}

\subsection{Main Results}
\textbf{Benchmarking CAD Code Generation.} Tab.~\ref{result} presents a comprehensive evaluation on CADExpert. The comparison includes nine state-of-the-art pretrained vision-language models (VLMs), all evaluated without any task-specific post training. We also benchmark two fine-tuned baselines: CAD-RL ~\cite{niu2025intent} and our method CME-CAD.
As illustrated in Fig.~\ref{3}, our dataset significantly outperforms existing open-source datasets in terms of complexity. A detailed comparison can be found in the supplementary materials. In this challenging context, our method achieves comprehensive superiority across all evaluation metrics. Specifically, we attain an IoU of 80.71\%, representing a performance improvement of 8.87\% points over the current SOTA methods. This demonstrates the effectiveness of our approach and its ability to handle more complex and realistic CAD design tasks. These improvements are also attributed to a significant increase in code executability, with our method achieving a code executability rate of 98.25\%. Such performance is highly valuable in real-world applications, as it means engineers can rely heavily on the model’s outputs, needing only to make minimal modifications to a small subset of more complex samples.

\begin{table}[t]
\centering
\small
\setlength{\tabcolsep}{2pt} 
\caption{Results of Multi-Expert learning compared to individual Expert learning.The table shows four sections: performance of a single expert with SFT, performance with multi-expert data in SFT, performance of a single expert with both SFT and GRPO, and the final results of our approach.}
\label{MoE-SFT}
\begin{tabular}{l|cccc}
\toprule
 {\textbf{\textsc{Model}}} 
& IOU(\%)$\uparrow$ & Mean CD$\downarrow$  & Med CD$\downarrow$ &Exec.(\%)$\uparrow$ \\

  \midrule
Expert1-SFT & 44.02 & 5.36 & 3.04 & 85.62 \\
Expert2-SFT & 38.08 & 7.05 & 3.74 & 82.14\\
Expert3-SFT & 40.91 & 5.34 & 3.27 & 85.35\\
  \midrule
MoE-SFT (Expert1) & 57.19 & 2.99 & 0.73 & 95.12\\
MoE-SFT (Expert2)& 61.30 & 2.23 & 0.63 &{97.06} \\
MoE-SFT (Expert3)& {64.45} & {1.63} & {0.42} & 96.99 \\
\midrule
E1-SFT-GRPO & 53.42 & 3.97 & 0.98 & 95.63\\
E2-SFT-GRPO & 49.29 & 4.74 & 1.51 & 93.96\\
E3-SFT-GRPO & 52.38 & 4.70 & 1.23 & 96.88 \\
  \midrule
  Ours (E1)& 75.69 & 1.28 & 0.28 & \textbf{98.35} \\
  Ours (E2)& 75.06 & 1.21 & 0.23 & 97.18\\
 Ours (E3)& \textbf{80.71} & \textbf{1.00} & \textbf{0.14} & 98.25 \\
\bottomrule
\end{tabular}
\end{table}

\begin{table}[t]
\centering
\small
\setlength{\tabcolsep}{4pt} 
\caption{Ablation studies on each component of CME-CAD, with results reported for the best-performing experts for brevity. ``EIAE'' refers to Expert-Internal Advantage Estimation, ``HSB'' stands for Hard Negative Sample Buffering Mechanism, and ``MECL'' denotes Multi-Expert Collaborative Learning.}

\label{aba}
\begin{tabular}{ccc|ccc}
\toprule
EIAE & HSB & MECL & IOU(\%)$\uparrow$   & Mean CD$\downarrow$ &Exec.(\%)$\uparrow$ \\
\midrule
\ding{55} &\ding{55} & \ding{55} & 64.45 & 1.63 & 96.99    \\

{\checkmark}& \ding{55} & \ding{55} & 73.89 & 1.33 & \textbf{98.28} \\

\checkmark  &\checkmark &\ding{55}  & 78.14 & 1.13 & 98.22 \\

\checkmark & \ding{55} & \checkmark & 76.71 & 1.28 & 98.19\\
\checkmark & \checkmark  &\checkmark& \textbf{80.71} &\textbf{1.00} & 98.25 \\
\bottomrule
\end{tabular}
\end{table}

\textbf{Ablation Study for Multi-Expert Learning and Individual Expert learning.}
Tab.~\ref{MoE-SFT} illustrates the performance differences between multi-expert learning and single-expert learning. As shown in the table, compared to using a single expert for SFT, multi-expert training leads to improvements across all evaluation metrics for each expert model. This demonstrates the successful facilitation of collaborative learning among the models. Despite completing the full SFT + GRPO process, a single expert model faces a performance ceiling due to the inherent limitations of its reasoning path. By incorporating data from multiple experts, the model can overcome some of these limitations with just SFT. Ultimately, applying our complete methodology results in significant improvements, surpassing the performance achieved by a single expert.

\textbf{Ablation Studies on Components of CME-CAD.} 
Tab.~\ref{aba} demonstrates the impact of each individual component of CME-CAD on the model's performance. Notably, the introduction of Expert-Internal Advantage Estimation leads to significant improvements in code executability. This is primarily due to its role in helping the model select the most suitable expert for each task. The Hard Negative Sample Buffering Mechanism has the most substantial effect on overall performance, which can be attributed to the high complexity of the CADExpert dataset. By continuously increasing the utilization of difficult samples during training, the model's performance is significantly enhanced. Furthermore, the joint use of Expert-Internal Advantage Estimation and Multi-Expert Collaborative Learning fully activates the potential of our training framework, enabling the model to effectively collaborative learning across different heterogeneous experts.

\section{Conclusion}
In this work, we introduce the CME-CAD paradigm, which leverages the complementary strengths of heterogeneous expert models to address key challenges in the generation of executable and editable CAD code. Our method effectively bridges the gap between the complexities of industrial design workflows and the capabilities of machine learning models, offering a novel solution to automate the creation of precise CAD models from 2D drawings input. By combining Multi-Expert Fine-Tuning (MEFT) and Multi-Expert Reinforcement Learning (MERL), we enable the model to explore diverse reasoning paths and improve performance, particularly in complex design scenarios where traditional methods fall short.
Furthermore, the introduction of CADExpert provides a high-quality benchmark that facilitates the development and evaluation of CAD code generation methods. This dataset, specifically designed to meet industrial-grade requirements, enables more robust training and evaluation. 
Overall, this work lays the foundation for more efficient, accurate, and editable CAD model generation.

{
    \small
    \bibliographystyle{ieeenat_fullname}
    \bibliography{main}
}


\end{document}